\documentclass[a4paper,fleqn]{cas-dc}

\usepackage[numbers,sort&compress]{natbib}
\usepackage{amsmath}
\usepackage{threeparttable}
\usepackage{soul} 
\usepackage{xcolor} 
\usepackage{subfigure}

\def\tsc#1{\csdef{#1}{\textsc{\lowercase{#1}}\xspace}}
\tsc{WGM}
\tsc{QE}

\begin{document}
\let\WriteBookmarks\relax
\def\floatpagepagefraction{1}
\def\textpagefraction{.001}

\shorttitle{}    

\shortauthors{}  

\title [mode = title]{Unveiling Large Language Models Generated Texts: A Multi-Level Fine-Grained Detection Framework}  




\author[1]{Zhen Tao}
\address[1]{School of Information, Renmin University of China, Beijing, China}

\author[2]{Zhiyu Li}
\address[2]{Institute for Advanced Algorithms Research, Shanghai, China}

\author[3]{Runyu Chen}
\address[3]{School of Information Technology and Management, University of International Business and Economics, China}

\author[4]{Dinghao Xi}[orcid=0000-0002-1685-8066]
\cormark[1]
\ead{xidinghao@mail.shufe.edu.cn}
\cortext[cor1]{Corresponding author}
\address[4]{Department of Digital Economics, Shanghai University of Finance and Economics, Shanghai, China}

\author[1]{Wei Xu}


\begin{abstract}
Large language models (LLMs) have transformed human writing by enhancing grammar correction, content expansion, and stylistic refinement. However, their widespread use raises concerns about authorship, originality, and ethics, even potentially threatening scholarly integrity. Existing detection methods, which mainly rely on single-feature analysis and binary classification, often fail to effectively identify LLM-generated text in academic contexts. To address these challenges, we propose a novel \textbf{M}ulti-level \textbf{F}ine-grained \textbf{D}etection (\textbf{MFD}) framework that detects LLM-generated text by integrating low-level structural, high-level semantic, and deep-level linguistic features, while conducting sentence-level evaluations of lexicon, grammar, and syntax for comprehensive analysis. To improve detection of subtle differences in LLM-generated text and enhance robustness against paraphrasing, we apply two mainstream evasion techniques to rewrite the text. These variations, along with original texts, are used to train a text encoder via contrastive learning, extracting high-level semantic features of sentence to boost detection generalization. Furthermore, we leverage advanced LLM to analyze the entire text and extract deep-level linguistic features, enhancing the model’s ability to capture complex patterns and nuances while effectively incorporating contextual information. Extensive experiments on public datasets show that the MFD model outperforms existing methods, achieving an MAE of 0.1346 and an accuracy of 88.56\%. Our research provides institutions and publishers with an effective mechanism to detect LLM-generated text, mitigating risks of compromised authorship. Educators and editors can use the model's predictions to refine verification and plagiarism prevention protocols, ensuring adherence to standards. 
\end{abstract}


\maketitle

\section{Introduction}\label{sec1}

The advent of large language models (LLMs) has revolutionized artificial intelligence, capturing widespread attention and transforming text generation across industries such as social media \cite{2023co}, recommendation system \cite{2023bookgptli}, medical diagnostics \cite{2024evaluation}, creative writing \cite{2024tree}, and intelligent education \cite{2023written}. While these advancements have unlocked significant benefits across various fields, they have also introduced potential risks and ethical concerns that warrant careful consideration \cite{2023science, 2023empowering}.

A key concern with LLM-generated content is the risk of misinformation, as these models, while capable of producing fluent and convincing text, can sometimes generate factually incorrect or misleading information \cite{2024bias, 2024exploring}. Another significant issue is the potential for LLMs to generate content that closely mimics original works, raising concerns about plagiarism and diminishing the value of authentic human creativity and professional expertise \cite{2023written, 2023overview}. In academia, these challenges are particularly concerning, as the distinction between human and machine-generated content becomes increasingly blurred. Issues surrounding authorship, originality, and the ethical use of LLMs in scholarly work are growing more complex \cite{2023written}. Given the paramount importance of accuracy, credibility, and originality in research, these risks threaten the integrity of academic work, underscoring the urgent need for more robust evaluation frameworks to safeguard scholarly standards \cite{2024practical}.

In response to this pressing need, researchers have developed two primary approaches for detecting LLM-generated text: \textbf{metric-based} and \textbf{model-based} algorithms \cite{2024llm}. Metric-based algorithms analyze linguistic features such as perplexity, word usage, and syntactic structures to detect potential signs of LLM involvement \cite{2024ACMsurvey}, using statistical measures to identify anomalies that suggest non-human authorship. Nevertheless, due to the continuous advancement of LLMs, the efficacy of these approaches is gradually diminishing \cite{2024authorship}. In contrast, model-based algorithms use machine learning models trained on large labeled datasets to detect subtle distinctions between LLM-generated and human-written content, often surpassing metric-based approaches \cite{2024authorship}. Model-based methods, which primarily extract high-level semantic features from text, have shown potential in detecting outputs from advanced models like ChatGPT. However, these methods often fail when applied to detecting content generated by the most sophisticated language models \cite{2023Frontiers}. While both metric-based and model-based approaches offer valuable insights, they face significant limitations when applied to the increasingly complex task of detecting LLM-generated text.

Firstly, most existing methods concentrate on a singular type of feature, such as statistical patterns or semantic content, rather than providing a comprehensive, multi-level analysis of textual features. This narrow focus often results in incomplete feature extraction, limiting the overall effectiveness of these detection methods. With LLM-generated text becoming increasingly similar to human-written text, single-level analysis is no longer sufficient, leading to both false positives and false negatives, thereby compromising the reliability of detection systems. Secondly, current models exhibit significant limitations in robustness \cite{2024navigating}. Even minor changes in wording or phrasing can produce markedly different detection results, particularly when researchers intentionally introduce subtle modifications like synonym substitutions or altered word order to evade detection \cite{2024hidding}. Enhancing model resilience to these disruptions is crucial for ensuring stable performance across diverse linguistic contexts. Furthermore, much of the existing research focuses on determining whether an entire text is LLM-generated, which lacks the precision required for academic integrity. Academic papers demand more detailed analysis, targeting specific sections or passages likely generated by LLMs. Without such granularity, detection methods may fail to effectively safeguard originality and credibility in scholarly work.

To address the aforementioned challenges, we propose a \textbf{M}ulti-level \textbf{F}ine-grained \textbf{D}etection (\textbf{MFD}) framework that integrates statistical, semantic, and linguistic features across different levels, specifically targeting text at the sentence level for more precise detection. The framework begins by extracting low-level statistical features, such as readability and authorstyle, to offer a quantitative view of the sentence’s structure. In parallel, high-level semantic features are captured using an encoder with contrastive learning, enabling the model to finely detect subtle semantic differences between LLM-generated and human-authored text. Additionally, advanced LLM are utilized to analyze the entire text, extracting deep linguistic features with a focus on lexicon, grammar, and syntax. This approach ensures that the model captures both sentence-level details and contextual information from the broader text. Notably, to counter common evasion tactics in LLM-generated text detection, we prompt the LLM to rewrite the original LLM text using two prevalent methods. By training the encoder with contrastive learning on both the original and the rewritten texts, we enhance the model's robustness in detecting unfamiliar or manipulated text. Furthermore, to comprehensively assess the degree of LLM involvement in text, our framework predicts each sentence of the academic text from the perspectives of lexicon, grammar, and syntax, enabling precise evaluations of LLM-generated text. Extensive experiments demonstrate the effectiveness of our framework, showing not only improved detection accuracy but also strong generalization across different types of LLM-generated text, thereby significantly contributing to the protection of academic integrity. Source data and coda are available at GitHub\footnote{https://github.com/TaoZhen1110/MFD}.

To summarize, our work makes four contributions:
\begin{itemize}
\item We propose a multi-level detection framework that integrates semantic, structural, and linguistic features to effectively detect LLM-generated text. 

\item Our framework predicts the degree of LLM involvement in each sentence by evaluating lexicon, grammar, and syntax, enabling a fine-grained analysis of LLM-generated text.

\item We improve model robustness by using contrastive learning on original and rewritten LLM text, generated through two common evasion tactics.

\item We leverage the advanced LLM to extract deep linguistic features, capturing both sentence-level details and broader contextual information.
\end{itemize}

The remainder of this paper is organized as follows. Section \ref{sec2} provides a comprehensive review of the related literature. Section \ref{sec3} elaborates on the detailed architecture and methodology of the proposed framework. Section \ref{sec4} outlines the experimental setup, followed by an in-depth analysis of the results and discussion. Finally, Section \ref{sec5} concludes the study by summarizing the key findings and highlighting avenues for future research.

\section{Related Work}\label{sec2}

\subsection{AI-generated Content Detection}

Large language models has revolutionized numerous industries with their exceptional text generation capabilities \cite{2024survey}. However, these advancements also pose risks of misuse. Overreliance on LLMs can facilitate the spread of misinformation \cite{2023canchen} and lead to unfair competition, threatening integrity in academic and societal contexts \cite{2023academicperkins, 2021algorithmicbirhane}. Therefore, timely detection of AI-generated text is crucial to mitigate these adverse impacts.

Existing detection methods are categorized into metric-based and model-based approaches. Metric-based methods quantify linguistic features, such as stylometric analysis and perplexity scoring, to determine text origin. Gehrmann et al. \cite{2019gltr} introduced GLTR, which uses metrics like word probability and entropy for detection, while Solaiman et al. \cite{2019release} proposed a zero-shot detection method leveraging pre-trained models like GPT-2 or GROVER. However, as LLMs advance, these methods have become less effective. To address these limitations, more advanced and sophisticated detection techniques have emerged. Mitchell et al. \cite{2023detectgpt} presented DetectGPT, a zero-shot method utilizing probability curvature analysis, and Tian et al. \cite{2023multiscale} developed the Multiscale Positive-Unlabeled (MPU) framework to enhance detection across varying text lengths. While metric-based approaches rely on linguistic analysis, model-based methods use advanced machine learning to classify text as human or AI-generated. These often involve fine-tuning language models or developing specialized architectures. For instance, Guo et al. \cite{2023close} fine-tuned a RoBERTa model to detect ChatGPT-generated text, while Wang et al. \cite{2023seqxgpt} introduced SeqXGPT, a sentence-level method using log-probability lists and self-attention networks for enhanced detection. Liu et al. \cite{2023coco} proposed COCO, which improves detection in low-resource settings by combining an entity coherence graph with contrastive learning. However, most methods focus on binary classification, which limits their application. As LLMs are frequently used for text expansion and refinement, fine-grained detection is increasingly necessary for practical use.

To address this gap, we propose a novel LLM involvement prediction algorithm. This algorithm quantifies LLM participation in text creation at the sentence level, identifying individual sentences that are likely LLM-generated. By moving beyond binary classification, our approach aims to ensure academic authenticity and uphold scholarly ethics, effectively tackling the challenges posed by modern text generation technologies.

\subsection{Multi-level Feature Learning}

Multi-level feature extraction and fusion are critical components in modern deep learning models, significantly enhancing performance in complex tasks, especially in text analysis \cite{2023multiscalelu}. These techniques involve extracting features at various levels of abstraction and combining them to create comprehensive data representations.

In the context of natural language processing (NLP), multi-level feature extraction captures both local and global linguistic patterns \cite{2023multiislam}. At lower levels, features may include lexical properties such as word embeddings and part-of-speech tags, providing foundational linguistic information \cite{2024parttian}. At higher levels, features encompass syntactic dependencies, semantic roles, and discourse structures, offering deeper insights into the text's meaning and context \cite{2023neuralkazanina}. By fusing these multi-level features, models develop a nuanced understanding of text, leading to improved performance in tasks such as sentiment analysis, machine translation, and information retrieval.

Applying multi-level feature learning is particularly pertinent for detecting LLM-generated text. Fine-grained detection requires analyzing textual data at various levels to identify subtle patterns indicative of LLM involvement. For instance, LLM-generated text may exhibit specific lexical choices, syntactic constructions, or discourse-level patterns that differ from human writing \cite{2023contrastingmunoz}. By leveraging multi-level features, our proposed algorithm can effectively capture these differences, enabling precise quantification of LLM involvement in LLM-generated text.

\subsection{Large Language Models and Thesis Writing}

Large language models like OpenAI's ChatGPT \cite{2023gptachiam} has transformed academic writing, extending beyond grammar correction to comprehensive support. While early tools like Grammarly and Hemingway Editor focused on linguistic accuracy, modern LLMs provide advanced capabilities, including text generation, structural suggestions, and literature review assistance \cite{2023exploring}. Studies indicate that these models alleviate writer's block, expedite manuscript preparation, and enhance paper quality \cite{2024using}. Research shows a growing reliance on LLMs in academic writing. Liang et al. \cite{2024mapping} analyzed 950,965 papers from arXiv and Nature (2020–2024), noting a consistent rise in LLM usage due to competitive pressures and academic challenges. Taiye et al. \cite{2024generative} highlighted generative LLM's potential to enhance student writing and critical thinking, advocating for integration through improved LLM literacy and regulation. While these models offer substantial benefits, they also have notable limitations. Safrai et al. \cite{2024utilizing} pointed out accuracy and reference reliability issues with LLM. Overreliance can erode critical thinking, lead to homogenized content, and diminish originality, thus impacting research quality and integrity \cite{2024chatgptmogavi, 2024llmszhang}.

These challenges highlight the need to monitor and manage LLM use in academic writing. To mitigate these risks, we propose a fine-grained detection framework that analyzes LLM involvement at the sentence level, enabling precise identification of LLM-generated text. This approach preserves academic integrity, ensures credibility, and encourages responsible use of LLMs in academia.

\section{The Multi-level Fine-grained Detection Framework}\label{sec3}

\begin{figure*}[t]
  \centering
  \includegraphics[width=16cm]{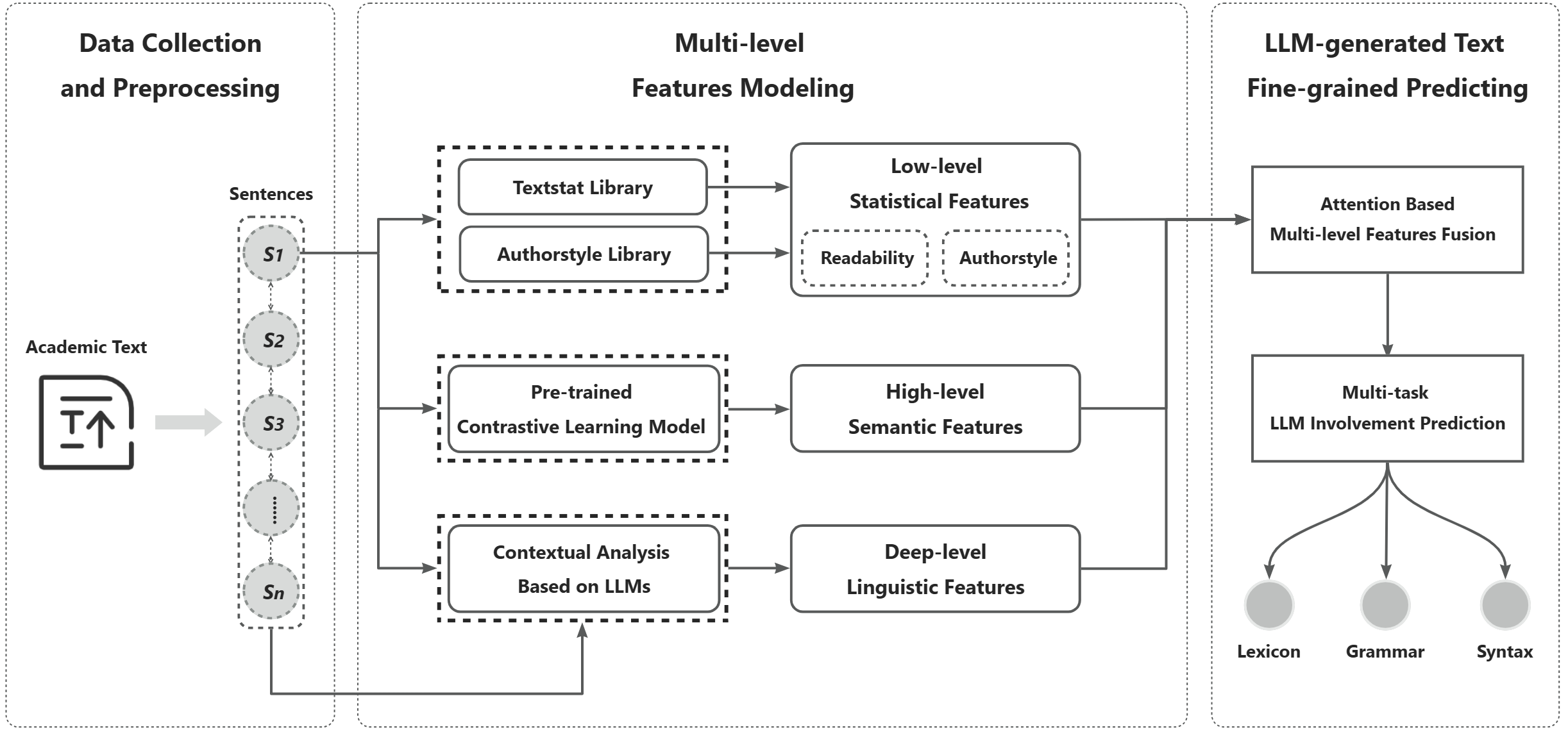}
   \caption{The Multi-level Fine-grained Detection (MFD) framework for LLM-generated text.}
   \label{fig1}
\end{figure*}

As illustrated in Fig. \ref{fig1}, our proposed framework begins by meticulously cleaning and preprocessing the academic text, segmenting it into individual sentences to enable focused analysis. Subsequently, we employ a suite of advanced methods to extract multi-level features from these sentences, ensuring that both sentence-level and broader contextual information are effectively captured. These features encompass statistical, semantic, and linguistic characteristics that are critical for discerning LLM-generated text. Finally, an attention-based fusion mechanism is utilized to integrate these diverse features, prioritizing the most informative signals. The fused representation is then fed into a dedicated LLM involvement predictor, designed for multi-task learning, to precisely evaluate LLM involvement across various linguistic dimensions.

\subsection{Problem Definition}

We propose the framework for sentence-level, fine-grained detection of LLM involvement in academic texts by assessing the degree of LLM involvement from three perspectives: lexicon, grammar, and syntax.

Given an academic text $D=\{s_{1}, s_{2}, ..., s_{n}\}$, where $s_{j}$ is the $j$-th sentence, the model outputs three distinct scores for each sentence $s_{j}$, representing the likelihood of LLM involvement in terms of lexical, grammatical, and syntactic features. The output for each sentence is:
\begin{equation}
\mathbf{r}_j = \left( r_{j}^{\text{lex}}, r_{j}^{\text{gram}}, r_{j}^{\text{syn}} \right),
\end{equation}
where $r_{j}^{\text{lex}}$ measures the probability of LLMs involvement in lexical choices, $r_{j}^{\text{gram}}$ represents the influence on grammatical structure, $r_{j}^{\text{syn}}$ captures syntactic patterns. For each sentence $s_{j}$, the probability distribution for these outputs is conditioned on the entire document context $D$:
\begin{equation}
\mathbf{r}_j = \mathbb{P}_k(r_j^{k} \mid s_j, D), \quad k \in \{\text{lex}, \text{gram}, \text{syn}\}
\end{equation}
Each component score $r_{j}^{\text{k}} \in [0,1]$, reflects the likelihood of LLM involvement from the respective perspective.

\subsection{Low-level Statistical Features}

To improve the detection of LLM-generated text at a structural level, we leverage low-level statistical features that offer quantitative insights into the text. These features focus on key linguistic metrics, particularly readability and authorstyle, serving as strong indicators of text originality and coherence.

\subsubsection{Readability}

Readability is a key feature for detecting anomalies in academic text, particularly in distinguishing human-authored content from LLM-generated text. Human writing often exhibits more variability in readability due to diverse sentence structures and vocabulary, while LLMs tend to produce text with uniform readability optimized for simplicity. As Patel et al. \cite{2024usepatel} demonstrated, LLMs like ChatGPT can reduce Flesch-Kincaid Grade (FKG) scores significantly, from 11.03 to 5.80 in patient education materials. While useful in educational contexts, such simplifications may indicate LLM authorship in academic writing, which typically requires greater complexity.

\begin{table}[b]
  \centering
  \caption{Detailed information on various readability numerical features.}
  \resizebox{\columnwidth}{!}{
    \begin{tabular}{lp{16.375em}}
    \toprule
    \textbf{Attribute} & \multicolumn{1}{l}{\textbf{Description}} \\
    \midrule
    Flesch Reading Ease & Measures how easy the text is to read, higher scores indicate easier readability \\
    Flesch Kincaid Grade & Indicates the U.S. school grade level needed to comprehend the text \\
    Smog Index & Calculates the complexity based on the number of polysyllabic words \\
    Coleman Liau Index & Assesses readability based on characters per word and sentence length \\
    Automated Readability Index & Uses character count and word length to estimate readability \\
    Dale Chall Readability Score & Compares text against a list of common words to measure readability difficulty \\
    Difficult Words & Counts the number of complex words  \\
    Linsear Write Formula & Evaluates readability based on the number of easy and difficult words \\
    Gunning Fog & Measures readability based on sentence length and complex word count \\
    Fernandez Huerta & A readability score based on sentence length and syllables per word \\
    Szigriszt Pazos & Calculates readability using a formula similar to Flesch indices \\
    Gutierrez Polini & A readability score designed for Spanish-language texts \\
    Crawford & Assesses readability by considering the proportion of complex words \\
    Gulpease & A readability measure  based on word and sentence length \\
    Osman & Another readability metric based on sentence structure and word length \\
    \bottomrule
    \end{tabular}}
  \label{tab1}%
\end{table}%

To capture differences in readability, we use various metrics from the textstat\footnote{https://github.com/textstat/textstat?tab=readme-ov-file} library (see Table \ref{tab1}). Key measures include Flesch Reading Ease and FKG, with LLM-generated text often scoring lower on FKG due to simpler vocabulary and sentence structures \cite{2024advancingjeong}. We also apply the Smog Index and Gunning Fog to assess sentence length and complex word usage, which tend to be more varied in human writing. Additional metrics, such as Coleman Liau and the Automated Readability Index, analyze character and sentence length to detect the oversimplifications typical of LLMs. The Dale Chall Score and Difficult Words Count further reveal LLMs' preference for simpler vocabulary, enhancing readability but diminishing academic depth \cite{2024andkhurana}. To ensure robustness, we include supplementary metrics like the Osman and Gulpease Index. These indices are useful for detection because they highlight the uniformity in LLM-generated text compared to the variability in human writing.

\subsubsection{Authorstyle}

Unlike readability metrics, which focus on surface-level complexity, authorstyle provides a deeper analysis of linguistic patterns and syntactic tendencies \cite{2024creatingguerin}. The authorstyle\footnote{https://github.com/mullerpeter/authorstyle} library extracts fine-grained features that distinguish human writing from LLM-generated text, particularly in academic contexts where stylistic consistency, complex structures, and lexical diversity signal expertise. By analyzing lexical richness, syntactic variation, and the use of punctuation and function words, it offers a comprehensive evaluation of writing quality beyond readability.

Table \ref{tab2} presents the various features of authorstyle. At the lexical level, metrics such as Average Word Length and Word Length Distribution reflect vocabulary choices, with LLM-generated text often using shorter words, while human writing shows more variability. Average Sentence Length (in words and characters) helps capture structural complexity, as human writing tends to be more intricate. Syntactic complexity is measured through Part-of-Speech (POS) Tag Frequency and Trigram Frequency, which highlight patterns in syntax and word usage that differ between human and LLM-generated text. Lexical diversity is assessed using Yule's K and Sichel's S, revealing vocabulary richness. Stylometric features, including Punctuation, Special Character, and Uppercase Frequency, along with function word and stopword ratios, further differentiate human writing from LLMs. N-gram Frequencies help detect repetitive phrases, often a hallmark of LLMs. Collectively, these features provide a comprehensive analysis of authorial style, enhancing the framework's ability to distinguish between human and LLM-generated text.

\begin{table}[b]
  \centering
  \caption{Detailed information on various authorstyle numerical features.}
  \resizebox{\columnwidth}{!}{
    \begin{tabular}{lp{14.71em}}
    \toprule
    \textbf{Attribute} & \multicolumn{1}{l}{\textbf{Description}} \\
    \midrule
    Average Word Length & Average number of characters per word \\
    POS Tag Frequency & Frequency of different part-of-speech tags  \\
    POS Tag Trigram Frequency & Frequency of three consecutive POS tags \\
    Word Length Distribution & Distribution of word lengths across the text \\
    Average Sentence Length Words & Average number of words per sentence \\
    Average Syllables Per Word & Average number of syllables per word \\
    Average Sentence Length Chars & Average number of characters per sentence \\
    Sentence Length Distribution & Distribution of sentence lengths to assess variability in sentence structure \\
    Yule K Metric & A measure of lexical diversity \\
    Sichel S Metric & Metric indicating the frequency of low-frequency words \\
    Average Word Frequency Class & The average word frequency class, indicating how often words are typically used in language. \\
    Punctuation Frequency & Frequency of punctuation marks \\
    Special Character Frequency & Frequency of punctuation marks \\
    Uppercase Frequency & Frequency of uppercase letters, capturing usage patterns such as acronyms or emphasis. \\
    Number Frequency & Frequency of numbers in the text \\
    Functionword Frequency & Frequency of function words \\
    Most Common Words Without Stopwords & The most frequently used content words  \\
    Stopword Ratio & Ratio of stopwords  to total words \\
    Top Word Bigram Frequency & Frequency of the most common bigrams \\
    Top Bigram Frequency & Frequency of bigrams, helping to detect repeated patterns of word combinations. \\
    Top 3 Gram Frequency & Frequency of trigrams (three consecutive words), indicating formulaic or common multi-word phrases. \\
    \bottomrule
    \end{tabular}}
  \label{tab2}%
\end{table}%

\subsection{High-level Semantic Features Based on Contrastive Learning}

\begin{figure*}[t]
  \centering
  \includegraphics[width=15cm]{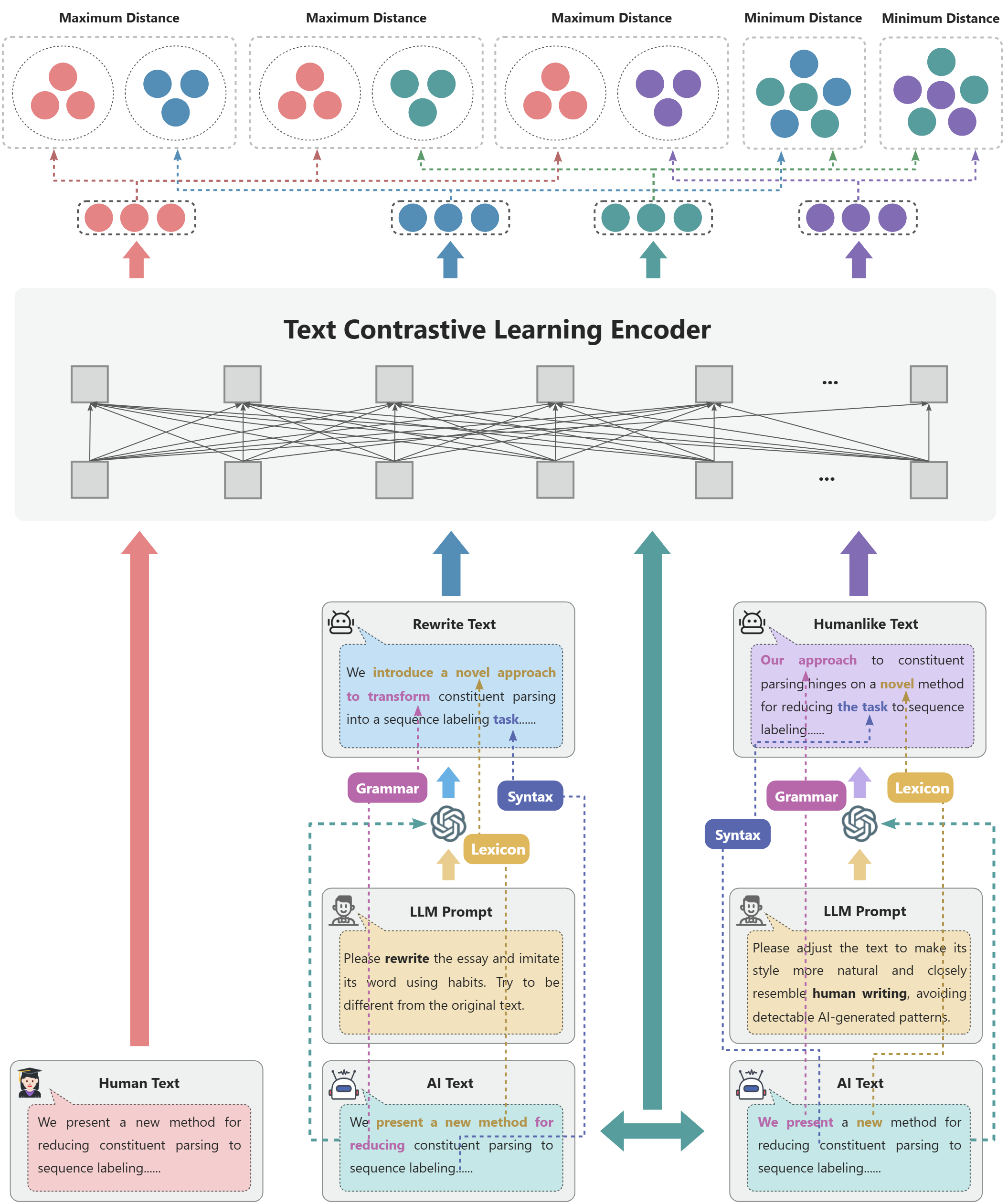}
   \caption{High-level semantic contrastive learning framework with adversarial training.}
   \label{fig2}
\end{figure*}

Existing detection systems struggle to distinguish between human and LLM-generated text, as LLMs can produce contextually accurate, coherent content that mimics human writing styles. Traditional detection methods based on superficial features are becoming less effective, especially when users modify or paraphrase LLM-generated text to evade detection. This complicates the distinction, as current approaches, particularly those relying on high-level semantic features, lack the robustness to capture these subtle changes.

To address these challenges, we propose a contrastive learning-based method with adversarial training to capture high-level semantic differences between human and LLM-generated text. The specific method diagram is shown in Figure \ref{fig2}. Specifically, we leverage an LLM to generate two additional variants of the LLM-generated text from the parallel dataset\footnote{https://github.com/FreedomIntelligence/ChatGPT-Detection-PR-HPPT/tree/main/Dataset}, focusing on common evasion tactics such as paraphrasing via LLMs \cite{2024outfoxkoike} and prompt-based adjustments \cite{2024stumbling}. Therefore, our method focuses on these prevalent tactics, creating variations of LLM-generated text that reflect real-world evasion techniques. First, leveraging the LLMs' text generation capabilities and extensive knowledge base, we design prompts that enable the LLM to closely imitate human writing patterns and expression habits:
\begin{equation}
T_{human-like} = LLM(P_{human}|T_{LLM}),
\end{equation}
where $T_{human-like}$ represents text that mimics human writing, $P_{human}$ is the prompt guiding this generation, and $T_{LLM}$ is the original LLM-generated text. Similarly, we create rewritten versions by prompting the LLM to paraphrase the original content while retaining its meaning:
\begin{equation}
T_{rewritten} = LLM(P_{rewrite}|T_{LLM}),
\end{equation}
where $T_{rewritten}$ denotes the paraphrased text, $P_{rewrite}$ is the paraphrasing prompt. These two variations are created because they align with common tactics used to evade detection.

After generating the human-like and rewritten variations of the LLM-generated text, we train the encoder using contrastive learning with four text types: the human-like text, the rewritten text, the original LLM-generated text, and the original human-authored text. This enables the model to capture fine-grained semantic differences between human and LLM-generated text, even after paraphrasing or modifications. We apply a \textbf{twice-triplet loss} function, where the original LLM-generated text serves as the anchor, the human-like and rewritten texts act as two positive examples, and a human-authored text from a different sample serves as the negative example, providing contrast for effective learning. To shift the cosine similarity values from $[-1, 1]$ to $[0, 1]$ and ensure non-negative distances, we adjust it by adding 1 and dividing by 2. The adjusted cosine similarity between any two text embeddings $T_i$ and $T_j$ is defined as:
\begin{equation}
\text{sim}^\ast(T_i, T_j) = \frac{T_i \cdot T_j}{2 \|T_i\| \|T_j\|} + \frac{1}{2}.
\end{equation}
The corresponding distances can be grouped into positive and negative examples, which are defined as:
\begin{eqnarray}
d_{\text{pos}} &=& 1 - \text{sim}^\ast(T_{\text{LLM}}, T_{\text{modified}}),
\\ \nonumber
d_{\text{neg}} &=& 1 - \text{sim}^\ast(T_x, T_{\text{human}}),
\end{eqnarray}
where \( T_{\text{modified}} \in \{T_{\text{human-like}}, T_{\text{rewritten}}\} \), representing the distance between the LLM-generated text and either the human-like or rewritten version of the text, \( T_x \in \{T_{\text{LLM}}, \\T_{\text{human-like}}, T_{\text{rewritten}}\} \), indicating the distance between the LLM-generated, human-like, or rewritten text and a genuine human-authored text. The twice-triplet loss function \(\mathcal{L}_{\text{contra}}\) is reformulated as:
\begin{equation}
\mathcal{L}_{\text{contra}} = \sum_{k=1}^{2} \sum_{l=1}^{3} \left( \left[ d_{\text{pos}}^k - d_{\text{neg}}^l + \alpha \right]_+ \right),
\end{equation}
where \([x]_+ = \max\{0, x\}\) denotes the hinge function, \(d_{\text{pos}}^k\) represents the distance between the anchor \(T_{\text{LLM}}\) and the \(k\)-th positive example (\(T_{\text{human-like}}\) for \(k=1\) and \(T_{\text{rewritten}}\) for \(k=2\)), \(d_{\text{neg}}^l\) denotes the distance between the anchor and the \(l\)-th negative example (\(T_{\text{LLM}}, T_{\text{human-like}}, T_{\text{rewritten}}\) compared with \(T_{\text{human}}\) for \(l = 1, 2, 3\)), and \(\alpha > 0\) is the margin parameter controlling the separation between positive and negative pairs.

By minimizing this loss function, the encoder learns to pull the anchor (LLM-generated text) closer to the positive examples (human-like and rewritten texts) while pushing it away from the negative example (human-authored text) in the embedding space. This process strengthens the model's ability to distinguish between human and LLM-generated text, even after paraphrasing or rewriting.

\begin{figure*}[t]
  \centering
  \includegraphics[width=15cm]{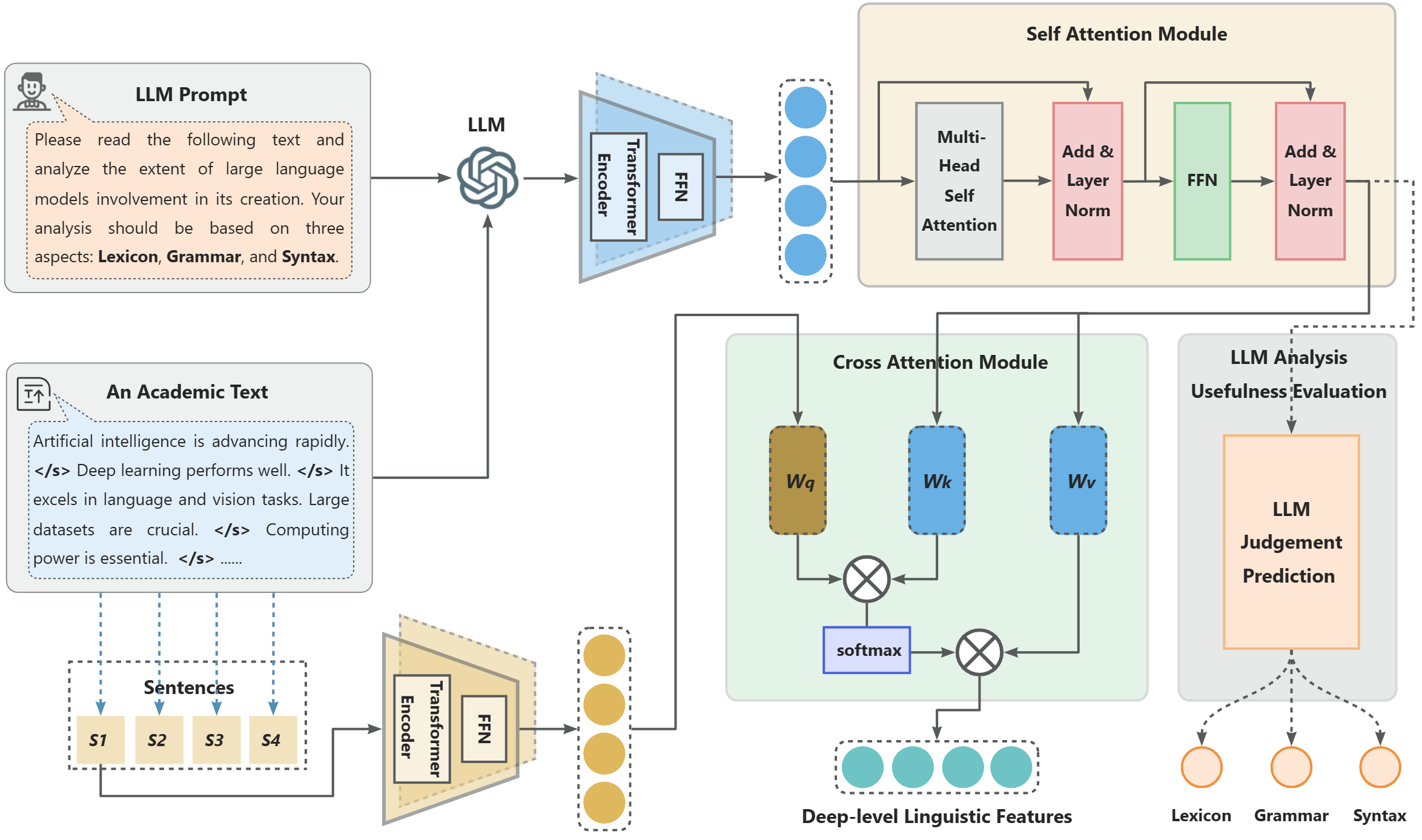}
   \caption{LLM-based deep linguistic feature extraction framework.}
   \label{fig3}
\end{figure*}

\subsection{Deep-level Linguistic Features Based on LLMs}

LLMs excel in natural language understanding, particularly in analyzing lexicon, grammar, and syntax \cite{2023largekaranikolas}. Their extensive pre-training enables them to capture complex linguistic patterns and subtleties that are often overlooked by traditional methods. By modeling probabilistic language structures, LLMs can detect deviations in word choice, syntax, and grammatical consistency—key indicators of LLM-generated text \cite{2024formohmer}. This enables LLMs to offer a refined approach to detecting their own involvement in text generation.

Our proposed model, illustrated in Figure \ref{fig3}, goes beyond sentence-level analysis by leveraging an LLM to extract deep linguistic features from the entire text, focusing on lexicon, grammar, and syntax. This holistic approach captures contextual relationships across the entire text, enabling the model to detect patterns in LLM-generated text that may be missed by superficial analysis. We begin by encoding both the original text and the LLM-analyzed text:
\begin{equation}
E_{i} = \text{Encoder}(T_{i}), \quad \text{where } i \in \{s_{j}, LLM\_analy\}.
\end{equation}
Here $s_{j}$ represents the $j$-th sentence of the entire text $T_{\text{orig}}$, which is encoded sentence by sentence, while $T_{\text{LLM\_analy}}$ refers to the entire text analyzed by the LLM. To refine the encoded features from the LLM analysis, we apply a Multi-Head Self-Attention (MHSA) mechanism followed by a Feed-Forward Network (FFN), similar to the Transformer architecture. This setup captures text dependencies and allows the model to focus on salient sections while using non-linear transformations for more expressive feature representation. For each input embedding $E_{LLM\_analy}$, we compute the queries $Q_{i}$, keys $K_{i}$, and values $V_{i}$ matrices for each attention head $i$:
\begin{equation}
X_{i} = E_{LLM\_analy} W_{i}^{X}, \quad X \in \{Q,K,V\},
\end{equation}
where $X_i$ represents any of the matrices $Q_i$, $K_i$, or $V_i$. $W_i^X$ corresponds to the learned weight matrices $W_i^Q$, $W_i^K$, or $W_i^V$ for the $i$-th attention head. The attention output for each head is computed as:
\begin{equation}
\text{head}_i = \text{softmax}\left( \frac{Q_i K_i^\top}{\sqrt{d_{1k}}} \right) V_i,
\end{equation}
with $d_{1k}$ representing the dimensionality of the key vectors. The outputs from all heads are concatenated and projected:
\begin{equation}
\text{MHSA}(E_{LLM\_analy}) = \text{Concat}(\text{head}_1, \dots, \text{head}_h) W^O,
\end{equation}
where $W^O$ is the output projection matrix. A residual connection and layer normalization are then applied to stabilize training:
\begin{equation}
E_{LLM\_analy}^{'} = \text{LN}(E_{\text{LLM\_analy}} + \text{MHSA}(E_{LLM\_analy})).
\end{equation}
The refined representation is further processed by a position-wise FFN, introducing non-linearity:
\begin{equation}
\text{FFN}(E_{LLM\_analy}^{'}) = \text{ReLU}(E_{LLM\_analy}^{'} W_1 + b_1) W_2 + b_2,
\end{equation}
where $W_{1}$ and $W_{2}$ are learned weight matrices, and $b_{1}$ and 
$b_{2}$ are bias vectors. Another residual connection and layer normalization yield the final representation:
\begin{equation}
E_{LLM\_analy}^{*} = \text{LN}(E_{LLM\_analy}^{'} + \text{FFN}(E_{LLM\_analy}^{'})).
\end{equation}
This streamlined architecture enhances the model's ability to learn complex linguistic patterns by retaining the input signal while applying transformative operations.

To extract deep linguistic features from individual sentences more effectively, while still allowing the model to account for contextual information, we introduce a Cross-Attention module. This mechanism facilitates interactions between the sentence and the LLM-analyzed text, ensuring that the analyzed text is more precise and contextually relevant to the sentence within its broader context. Let the feature representation of the sentence be $Q = E_{s_{j}}$ and that of the LLM-analyzed text be $K = V = E_{LLM\_analy}^{*}$. The Cross-Attention mechanism then computes the standard scaled dot-product between the normalized query and key vectors, leveraging the interactions between the sentence and LLM-analyzed text. This produces attention scores that highlight specific aspects of the LLM-analyzed text, making it more focused on this thesis sentence:
\begin{equation}
\text{Att}_{\text{scores}} = \frac{Q K^{\top}}{\sqrt{d_{2k}}},
\end{equation}
where \( d_{2k} \) is the dimensionality of the key vectors used as a scaling factor. The attention scores are then used to weight the value vectors \( V \), focusing on the most contextually relevant parts of the LLM-analyzed text:
\begin{equation}
f_{deep} = \text{softmax}\left( \text{Att}_{\text{scores}} \right) V.
\end{equation}
where \( f_{deep} \) represents the deep linguistic features of the sentence.

\subsection{Multi-level Features Fusion and LLM Involvement Prediction}

After extracting features from the three hierarchical levels, we perform a weighted fusion to integrate information from each level. This fusion is crucial for capturing both global contextual nuances and fine-grained textual characteristics. We introduce learnable weights $w_i$, applied to the respective feature vectors \( f_i \) corresponding to the low-level, high-level, and deep-level features:
\begin{equation}
\tilde{f}_i = w_i \odot f_i, \quad \text{for } i \in \{low, high, deep\},
\end{equation}
where \( \odot \) denotes element-wise multiplication. The weighted feature vectors \( \tilde{f}_i \) are then concatenated to form a unified representation:
\begin{equation}
\tilde{f}_{\text{fusion}} = \text{Concat}(\tilde{f}_{low}, \tilde{f}_{high}, \tilde{f}_{deep}),
\end{equation}
with $\text{Concat}$ indicating concatenation along the feature dimension. The fused vector \( \tilde{f}_{\text{fusion}} \) is subsequently fed into a Multi-Layer Perceptron (MLP) classifier to produce the final prediction. The prediction loss is calculated using the Mean Squared Error (MSE) between the model's output and the ground truth labels \( y \):
\begin{equation}
\mathcal{L}_{\text{Pred}} = \text{MSE}\left( \text{MLP}\left( \tilde{f}_{\text{fusion}} \right), y \right).
\end{equation}
As shown in Figure \ref{fig3}, to further assess the utility of the LLM-analyzed text, we input its refined representation \( E_{\text{LLM\_analy}}^{*} \) into an evaluator network, computing an additional loss term:
\begin{equation}
\mathcal{L}_{\text{LLM}} = \text{MSE}\left( \text{MLP}\left( E_{\text{LLM\_analy}}^{*} \right), y \right).
\end{equation}
The overall loss function is formulated as a weighted sum of the prediction loss and the LLM evaluation loss:
\begin{equation}
\mathcal{L} = \mathcal{L}_{\text{Pred}} + \beta \mathcal{L}_{\text{LLM}},
\end{equation}
where \( \beta \) is a hyperparameter that balances the contributions of the two loss terms. By jointly optimizing these objectives, the model effectively integrates multi-level features while leveraging insights from LLM analysis, thereby enhancing predictive performance.

\section{Empirical Analysis}\label{sec4}

\subsection{Dataset Description}

In this study, we use the \textbf{Paraphrased Text Span Detection}\footnote{https://huggingface.co/datasets/linzw/PASTED/blob/main/regression-multi-dimension.zip} (PASTED) dataset as the primary dataset for both training and testing, while \textbf{Multi-Paraphrase}\footnote{https://huggingface.co/datasets/linzw/PASTED/blob/main/eval-multi-paraphrase.zip} (MP) and \textbf{Out of Distribution-GPT4}\footnote{https://huggingface.co/datasets/linzw/PASTED/blob/main/eval-OOD.zip} (OOD-GPT4) are used as test datasets to evaluate the model's generalization capability. The PASTED dataset is large and provides continuous regression labels ranging from 0 to 1, making it ideal for a comprehensive assessment of the model's performance in regression tasks. In contrast, the MP and OOD-GPT4 datasets consist of binary classification labels (0 or 1) and are used exclusively for testing the model's generalization in classification tasks. This setup allows us to systematically evaluate the model's robustness and generalization across different task types and data distributions.

\textbf{PASTED}: This dataset aims to detect sentence-level LLM involvement in text generation of academic writing \cite{2024spottingli}. It is annotated with three labels—lexical, syntactic, and grammatical—to assess LLM involvement. From the dataset, 30,000 text samples were randomly selected and split into training, validation, and test sets (80\%, 10\%, 10\%). It includes content from models like GPT-3.5 and Dipper, with sentence-level segmentation using the '</s>' delimiter for precise LLM detection. 

\textbf{MP}: This dataset comprises 1,907 texts, each of which has been rewritten using contemporary LLMs, such as GLM. The labels for each sentence are also based on three dimensions: lexical, syntactic, and grammatical. Unlike the PASTED dataset, the MP dataset uses binary labels (0 or 1) for each of these dimensions. Due to the nature of these classification labels, we exclusively employ the dataset for testing purposes, aiming to evaluate the model's ability to generalize across distinct linguistic dimensions in classification tasks. 

\textbf{OOD-GPT4}: This dataset contains 9,386 texts, which were edited or rewritten using the latest GPT-4 model. In other aspects, it is essentially similar to the MP dataset. Therefore, we also used it as a test dataset.

\subsection{Evaluation Metrics}

We choose four common evaluation metrics for our proposed model: Mean Absolute Error (MAE), Mean Squared Error (MSE), Root Mean Squared Error (RMSE), and Accuracy. Since the task is a regression problem, accuracy is calculated by converting the predictions and true labels into binary form with a threshold of 0.5. The metrics—MAE, MSE, RMSE, and Accuracy—are computed for each dimension (lexical, syntactic, and semantic), and the final scores are obtained by averaging the results across these dimensions, providing a comprehensive evaluation of the model’s overall performance.

\subsection{Implementation Details}

In our experiments, we utilize the AdamW optimizer with a learning rate of \(1 \times 10^{-3}\) and a weight decay of \(1 \times 10^{-4}\). To further enhance the training process, we apply a StepLR learning rate scheduler with a step size of 5 and a decay factor \(\gamma = 0.5\). The model is trained for 30 epochs, with a training batch size of 512 and a validation batch size of 64. In this experiment, the value of $\beta$ in the loss function is set to 0.5, and the number of heads in the multi-head attention mechanism is set to 8. The encoder in our framework is based on T5 model, while the contrastive learning mechanism, which helps counter evasion tactics in text generation, as well as the deep linguistic feature extraction, both rely on the Llama-3.1-8B-Instruct model. Only parameters that require gradients are updated during optimization, ensuring computational efficiency. All models are trained and evaluated on four NVIDIA H800 GPUs.

\subsection{Experimental Results}

\subsubsection{Comparison with State-of-the-Art Detection Models}

To evaluate the effectiveness of our proposed method, we conduct a comparative analysis against several state-of-the-art models, each with a unique underlying mechanism. \textbf{GLTR} \cite{2019gltr} uses statistical analysis to visualize token-level distribution probabilities, aiding human detection of generated text. \textbf{SeqXGPT} \cite{2023seqxgpt} combines word-level log-probabilities with convolutional neural networks and attention mechanisms for sentence-level detection, excelling in mixed content scenarios. \textbf{Sniffer} \cite{li2023origin} focuses on perplexity features, particularly effective in low-resource cases. \textbf{PPL} \cite{2023close} uses perplexity scores from GPT models to distinguish machine-generated content. \textbf{PTD} \cite{2024spottingli} detects paraphrased spans, identifying subtle LLM-generated manipulations within human text. Lastly, \textbf{AdaLoc} \cite{2024machinezhang} employs multi-sentence predictions to localize AI-generated segments, enhancing detection precision.

\begin{table}[htbp]
  \centering
  \caption{Quantitative results of different models on key metrics.}
  \resizebox{\columnwidth}{!}{
    \begin{tabular}{p{11.46em}llll}
    \toprule
    \textbf{Model} & \multicolumn{1}{p{4.04em}}{\textbf{MAE}} & \multicolumn{1}{p{4.04em}}{\textbf{MSE}} & \multicolumn{1}{p{4.04em}}{\textbf{RMSE}} & \multicolumn{1}{p{4.04em}}{\textbf{Accuracy}} \\
    \midrule
    GLTR \cite{2019gltr}  & 0.2022  & 0.0773  & 0.2719  & 0.8686  \\
    SeqXGPT \cite{2023seqxgpt} & 0.2109  & 0.0772  & 0.2720  & 0.8687  \\
    Sniffer \cite{li2023origin} & 0.1989  & 0.0781  & 0.2734  & 0.8689  \\
    PPL \cite{2023close}  & 0.1807  & 0.0723  & 0.2633  & 0.8703  \\
    PTD \cite{2024spottingli}  & 0.1904  & 0.0765  & 0.2706  & 0.8687  \\
    AdaLoc \cite{2024machinezhang} & 0.1980  & 0.0770  & 0.2716  & 0.8686  \\
    \textbf{Proposed Model} & \textbf{0.1347} & \textbf{0.0612} & \textbf{0.2428} & \textbf{0.8856} \\
    \bottomrule
    \end{tabular}}
  \label{tab3}
\end{table}

The results of our comparative analysis, shown in Table \ref{tab3}, demonstrate the superiority of our proposed model across multiple evaluation metrics. In terms of MAE, our model achieves a significantly lower value of 0.1347, outperforming all other models, with the second best being PPL at 0.1807. Similarly, in terms of MSE and RMSE, our method reports the lowest values at 0.0612 and 0.2428, respectively, again surpassing PPL, which performs second best with MSE and RMSE scores of 0.0723 and 0.2633, respectively. Furthermore, our model achieves the highest accuracy of 0.8736, indicating a noticeable improvement over existing models such as PPL (0.8703) and AdaLoc (0.8686). These results highlight the effectiveness of our approach, particularly in accurately detecting LLM-generated text and minimizing prediction errors. The consistent outperformance across all metrics suggests that our model not only improves overall detection accuracy but also reduces errors more effectively compared to existing state-of-the-art methods.

\subsubsection{Generalization Performance Study}

\begin{figure*}[t]
\centering
\subfigure[MP]{
\label{fig4.1}
\includegraphics[width=0.49\textwidth]{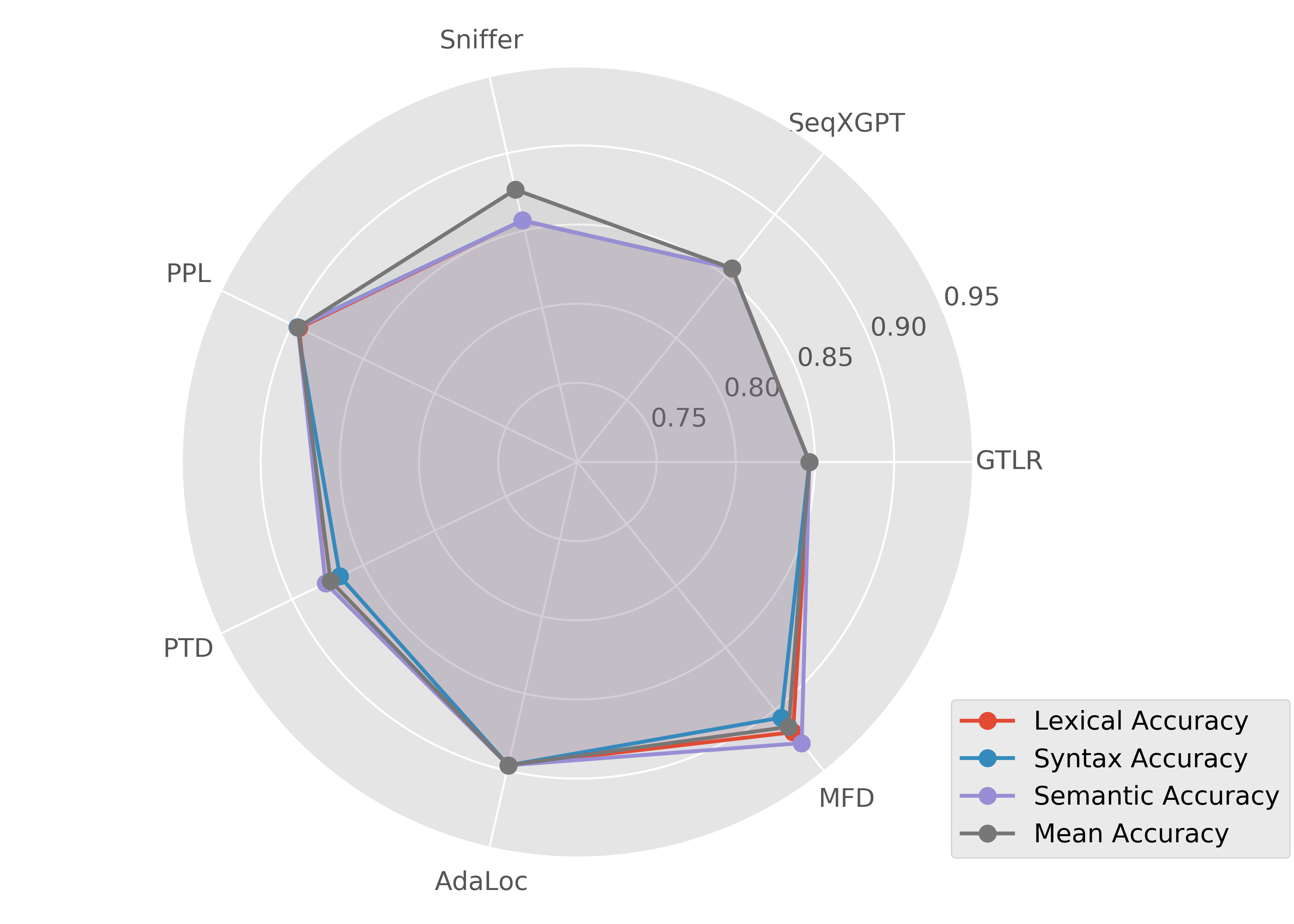}}
\subfigure[OOD-GPT4]{
\label{fig4.2}
\includegraphics[width=0.49\textwidth]{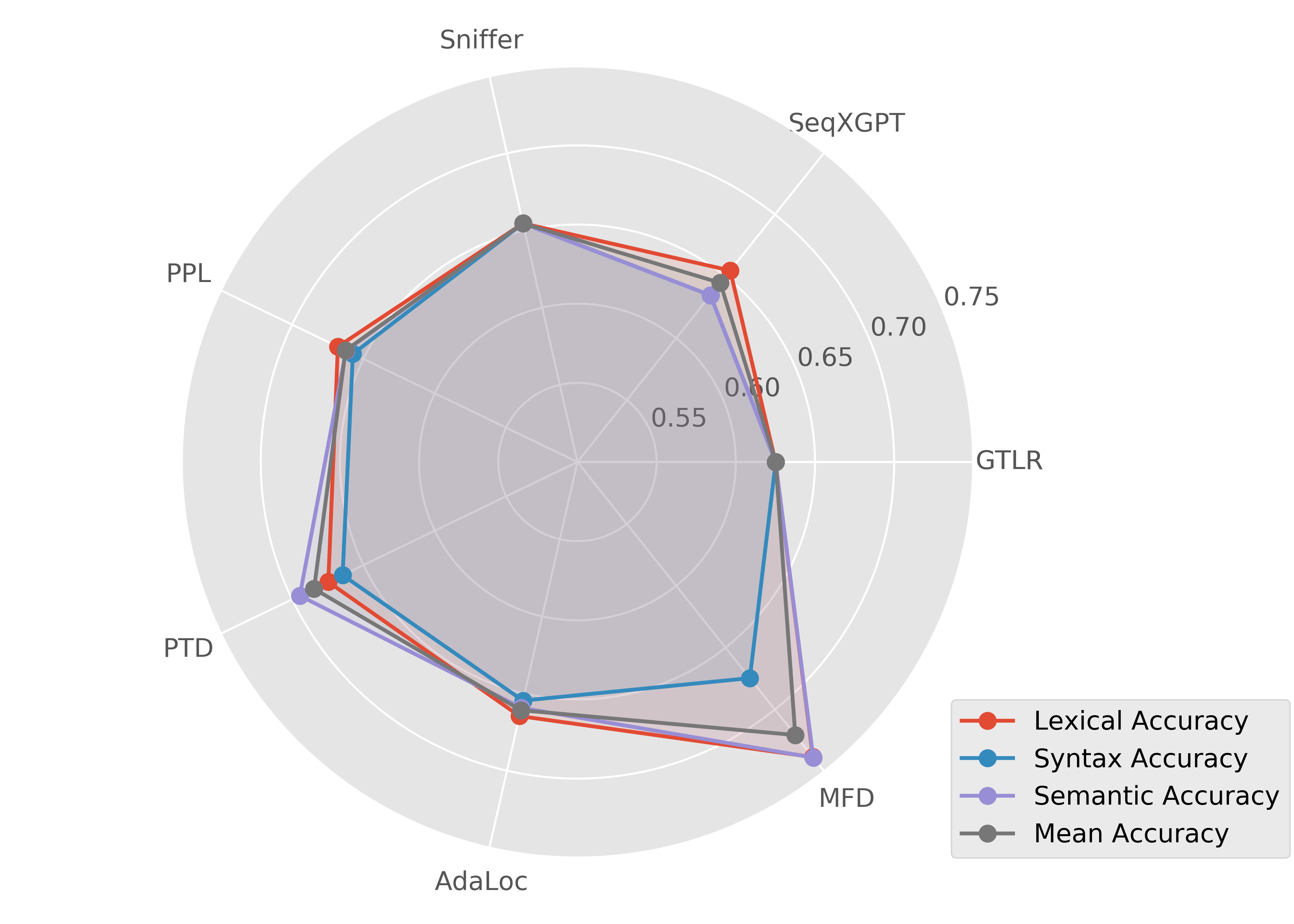}}
\caption{Radar charts of generalization performance of models on MP and OOD-GPT4 datasets.}
\label{fig4}
\end{figure*}

In this experiment, we evaluate the generalization performance of state-of-the-art models on the MP and OOD-GPT4 datasets, as depicted in Figure \ref{fig4}. The radar charts assess four dimensions: Lexical Accuracy, Syntax Accuracy, Semantic Accuracy, and Mean Accuracy. On the MP dataset, our proposed MFD model achieves the highest scores, particularly in Lexical and Semantic Accuracy, reflecting its robustness and strong detection capabilities. Even in the OOD-GPT4 dataset, which consists of text generated by the highly sophisticated GPT-4, MFD maintains superior performance across all dimensions, further emphasizing its strong generalization ability in the face of advanced language generation systems. Other models, such as AdaLoc and PPL, perform well on MP but exhibit noticeable drops in certain dimensions on OOD-GPT4, while GTLR and SeqXGPT consistently underperform. This highlights MFD's robustness and effectiveness in both standard and out-of-distribution scenarios.

\subsubsection{Ablation Study on Multi-level Feature Contributions}

The ablation study results in Table \ref{tab4} demonstrate the critical role of combining multi-level features in enhancing model performance. Individually, the low-level, deep-level, and high-level features show moderate performance, with accuracies ranging from 0.8629 to 0.8748. However, when these features are combined, particularly the low-level and high-level features, we observe a significant improvement, with an accuracy reaching 0.8837. The proposed model, which integrates all three levels of features, achieves the best performance across all metrics, with an MAE of 0.1347, an MSE of 0.0612, and an accuracy of 0.8856. This suggests that each feature set captures complementary information, and their integration enables the model to generalize better by leveraging both shallow patterns and deeper semantic cues within the data.

\begin{table}[htbp]
  \centering
  \caption{Results of ablation study on multi-level features.}
  \resizebox{\columnwidth}{!}{
    \begin{tabular}{lllll}
    \toprule
    \textbf{Model} & \textbf{MAE} & \textbf{MSE} & \textbf{RMSE} & \textbf{Accuracy} \\
    \midrule
    Low-level & 0.1929  & 0.0784  & 0.2799  & 0.8629  \\
    High-level & 0.1722  & 0.0662  & 0.2572  & 0.8748  \\
    Deep-level & 0.1891  & 0.0797  & 0.2762  & 0.8687  \\
    Low-level+High-level & 0.1387  & 0.0621  & 0.2444  & 0.8837  \\
    Low-level+Deep-level & 0.1624  & 0.0710  & 0.2609  & 0.8723  \\
    Deep-level+High-level & 0.1421  & 0.0626  & 0.2435  & 0.8839  \\
    \textbf{Proposed Model} & \textbf{0.1347} & \textbf{0.0612} & \textbf{0.2428} & \textbf{0.8856} \\
    \bottomrule
    \end{tabular}}
  \label{tab4}%
\end{table}%

\subsubsection{Ablation Study on Different LLMs}

In this experiment, we integrate various large language models into our framework to evaluate how different architectures impact performance. We test six state-of-the-art LLMs, each representing different design philosophies. \textbf{GLM-4-9B-Chat}, developed by Tsinghua University and Zhipu AI, is optimized for multi-turn dialogues, while \textbf{Baichuan2-13B-Chat}, from Baichuan Intelligence, focuses on multilingual tasks across diverse linguistic contexts. \textbf{InternLM2.5-7B-Chat}, designed by Shanghai AI Lab, is a lightweight model that balances efficiency with performance. \textbf{Yi-1.5-9B-Chat}, developed by 01.AI, is a fine-tuned model that excels in tasks requiring reasoning, commonsense understanding, and instruction-following capabilities. \textbf{Qwen-2.5-7B-Chat}, developed by Alibaba, offers fine-grained control in conversational interactions. Finally, \textbf{Llama-3.1-8B-Instruct}, from Meta AI, is fine-tuned for instruction-following, excelling in generating precise and adaptive responses.

The results in Table \ref{tab5} reveal that the performance differences among the tested LLMs are relatively small, with all models demonstrating a high level of capability due to the maturity of transformer-based architectures and extensive pretraining on large datasets. While InternLM2.5-7B-Chat slightly outperforms others in MAE (0.1316) and Qwen-2.5-7B-Chat edges out in accuracy (0.8857), Llama-3.1-8B-Instruct offers the most balanced performance, excelling across RMSE, MSE, and accuracy metrics. These subtle variations likely stem from differences in fine-tuning methods and data diversity, indicating that modern LLMs are converging in performance but can still be tailored for specific tasks to optimize results.

\begin{table}[htbp]
  \centering
  \caption{Results of ablation study on different LLMs.}
  \resizebox{\columnwidth}{!}{
    \begin{tabular}{lllll}
    \toprule
    \textbf{LLM} & \textbf{MAE} & \textbf{MSE} & \textbf{RMSE} & \textbf{Accuracy} \\
    \midrule
    GLM-4-9B-Chat & 0.1406  & 0.0616  & 0.2435  & 0.8850  \\
    Baichuan2-13B-Chat & 0.1442  & 0.0624  & 0.2431  & 0.8850  \\
    InternLM2.5-7B-Chat & \textbf{0.1316} & 0.0614  & 0.2431  & 0.8852  \\
    Yi-1.5-9B-Chat & 0.1423  & 0.0616  & 0.2435  & 0.8852  \\
    Qwen-2.5-7B-Chat & 0.1404  & 0.0614  & 0.2432  & \textbf{0.8857} \\
    \textbf{Llama-3.1-8B-Instruct} & 0.1347  & \textbf{0.0612} & \textbf{0.2428} & 0.8856  \\
    \bottomrule
    \end{tabular}}
  \label{tab5}%
\end{table}%

\subsubsection{Ablation Study on Different Encoders}

\begin{table}[htbp]
  \centering
  \caption{Results of ablation study on different encoders.}
  \resizebox{\columnwidth}{!}{
    \begin{tabular}{p{5.5em}llll}
    \toprule
    \textbf{Encoder} & \multicolumn{1}{p{5.125em}}{\textbf{MAE}} & \multicolumn{1}{p{5.75em}}{\textbf{MSE}} & \multicolumn{1}{p{5.04em}}{\textbf{RMSE}} & \multicolumn{1}{p{6.04em}}{\textbf{Accuracy}} \\
    \midrule
    Electra  & 0.1681  & 0.0696  & 0.2585  & 0.8733  \\
    XLNet  & 0.1711  & 0.0696  & 0.2585  & 0.8735  \\
    DistilBERT  & 0.1585  & 0.0672  & 0.2540  & 0.8757  \\
    ALBERT  & 0.1497  & 0.0792  & 0.2773  & 0.8687  \\
    RoBERTa  & 0.1651  & 0.0695  & 0.2583  & 0.8736  \\
    \textbf{T5} & \textbf{0.1347} & \textbf{0.0612} & \textbf{0.2428} & \textbf{0.8856} \\
    \bottomrule
    \end{tabular}}
  \label{tab6}%
\end{table}%

Table \ref{tab6} presents the performance of our framework using different contrastive learning encoders. The results show that T5 outperforms all other encoders across every metric, with the lowest MAE (0.1347) and the highest accuracy (0.8856). This superior performance is likely due to T5's versatile sequence-to-sequence architecture, which excels in a variety of tasks by effectively capturing both input and output relationships. In contrast, ALBERT shows the weakest results, particularly in RMSE (0.2773), likely due to its parameter-sharing mechanism, which may reduce model expressiveness. DistilBERT and RoBERTa perform competitively, demonstrating that compression (in DistilBERT’s case) and enhanced training (in RoBERTa’s case) can still yield high accuracy. Overall, these results highlight that while all encoders perform well, T5's architecture makes it particularly well-suited for tasks requiring nuanced predictions and lower errors.

\subsubsection{Ablation Study on Different Accuracy Threshold}

\begin{figure}[t]
  \centering
  \includegraphics[width=\columnwidth]{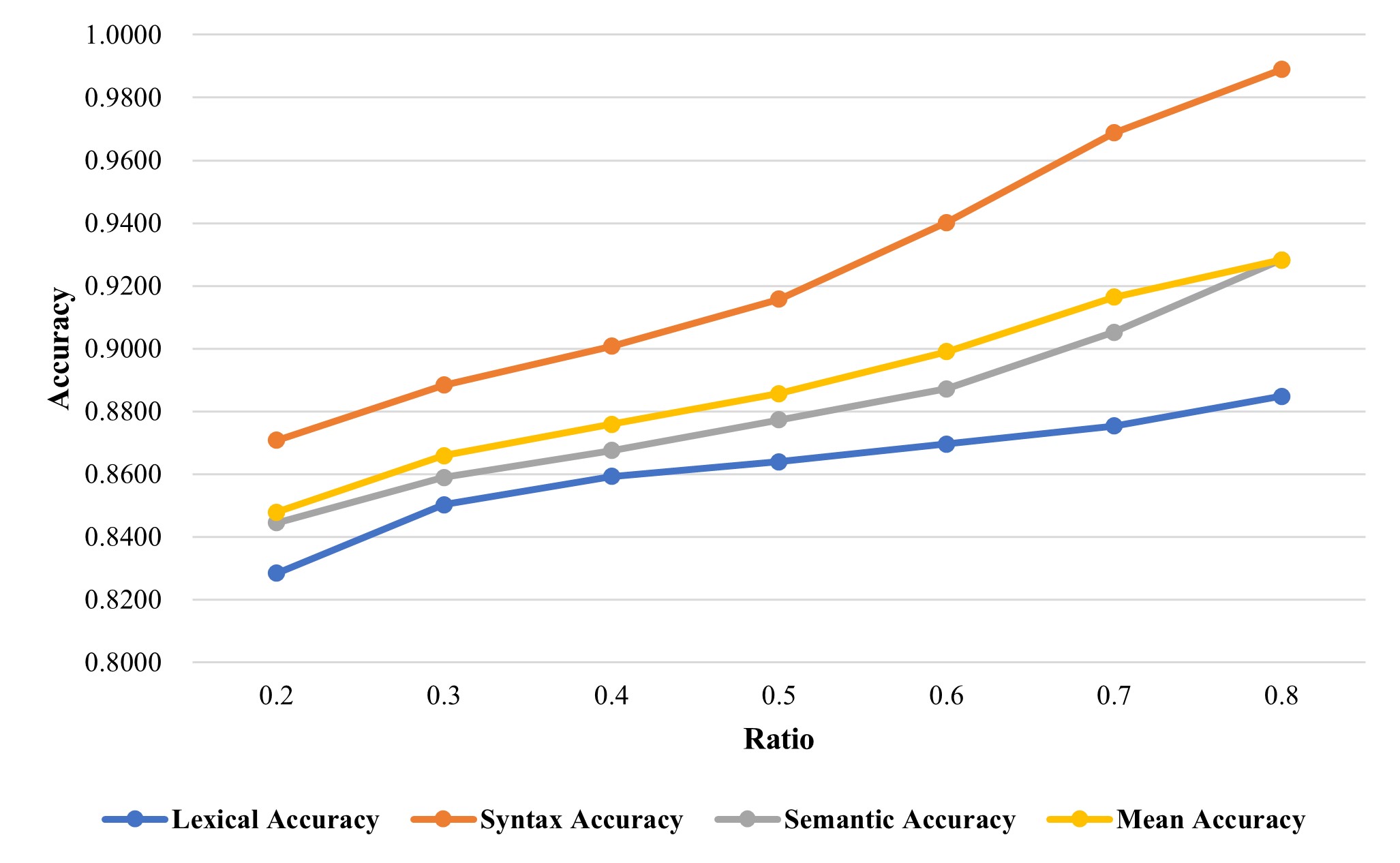}
   \caption{Effect of threshold ratios on Accuracy metrics.}
   \label{fig5}
\end{figure}

Since our predictions are continuous values between 0 and 1, we convert them into a classification task by setting thresholds to compute accuracy metrics. Figure \ref{fig5} demonstrates the effect of varying accuracy thresholds on different linguistic levels, including lexical, syntactic, and semantic accuracy. As the threshold ratio increases, all accuracy metrics improve, with the highest mean accuracy achieved at a ratio of 0.8. Higher thresholds force the model to make more confident predictions, effectively filtering out noise and focusing on more clearly defined patterns, particularly in syntactic and semantic dimensions. However, the improvements in lexical accuracy are more modest, indicating that lexical features are easier to capture even at lower thresholds. The chosen threshold of 0.5 in this study represents a balanced trade-off, offering competitive performance across all metrics while avoiding potential overfitting that may arise with higher thresholds. At this ratio, the model achieves a solid mean accuracy and ensures stable generalization without overly restricting the classification boundaries. This balance between precision and flexibility makes 0.5 an optimal choice for the task.

\subsubsection{Case Study of the MFD Model}

To demonstrate the effectiveness of our MFD model, we analyze two academic text samples to assess LLM involvement at the sentence level, as shown in Table \ref{tab7}. In the first example, the MFD model's predictions largely align with the true labels. However, there is a slight overestimation in the sixth and final sentences, where the predicted LLM involvement is deeper than the true labels. Despite this minor discrepancy, the model effectively captures the overall LLM-generated content in the text. In the second example, the model's predictions closely match the ground truth, with predicted LLM involvement falling within the expected range across almost all sentences, except for the second sentence, where the predicted LLM involvement is shallower than the true value. These case studies illustrate the MFD model's strength in fine-grained, sentence-level analysis, providing a reliable and nuanced assessment of LLM-generated text.

\begin{table*}[htbp]
  \centering
  \caption{Comparison of raw text and MFD detection results for LLM Involvement.}
  \renewcommand{\arraystretch}{1.5}
  \begin{threeparttable}
  \resizebox{\textwidth}{!}{
    \begin{tabular}{p{57.625em}}
    \toprule
    \textbf{Raw Text}: We present a class of efficient models called MobileNets for mobile and embedded vision applications. 
    MobileNets are based on a streamlined architecture that uses depthwise separable convolutions to build light sensitive images in the presence of occlusions, illumination changes or complex backgrounds. 
    \sethlcolor{cyan!40}
    \hl{The proposed models can be used in situations where small size is important, such as wearable devices and vehicular sensing systems.} 
    \sethlcolor{cyan!15}
    \hl{To demonstrate their performance, we test MobileNet V4 on several datasets including i-Sight 3D, CycleGANs, and AlexNet.}
    \sethlcolor{cyan!100}
    \hl{We show how our model can be used with these large scale video sequences on different platforms like smartphones.} 
    \sethlcolor{cyan!60}
    \hl{Our paper presents an efficient framework for creating high resolution stereo pictures from 2D videos using only image-level information.}
    \sethlcolor{cyan!60}
    \hl{Even in challenging conditions like low frame rate, poor lighting, and occlusion propagation, our approach can produce detailed views.} 
    \sethlcolor{cyan!60}
    \hl{Additionally, our method minimizes computational cost and reduces noise during preprocessing to maintain high quality.}
    \sethlcolor{cyan!40}
    \hl{It allows for regular interval image fitting without significant loss in accuracy.}
    \sethlcolor{cyan!40}
    \hl{It also provides real-time camera pose updates, as demonstrated by conducting human arm position estimation over a webcam stream with accuracies comparable to other methods.}
    \sethlcolor{cyan!15}
    \hl{In addition, full colour images were produced within 60 ms on device with limited processing power e.g., Galaxy S5 smartphone equipped with Adreno 320 GPU.} Finally most importantly our technique achieves good performance in terms of energy efficiency allowing to use a single thread of CPU with minimal memory requirement. \\
    \midrule
    \textbf{MFD Detection Result}: We present a class of efficient models called MobileNets for mobile and embedded vision applications. 
    MobileNets are based on a streamlined architecture that uses depthwise separable convolutions to build light sensitive images in the presence of occlusions, illumination changes or complex backgrounds. 
    \sethlcolor{pink!40}
    \hl{The proposed models can be used in situations where small size is important, such as wearable devices and vehicular sensing systems.}
    \sethlcolor{pink!15}
    \hl{To demonstrate their performance, we test MobileNet V4 on several datasets including i-Sight 3D, CycleGANs, and AlexNet.}
    \sethlcolor{pink!100}
    \hl{We show how our model can be used with these large scale video sequences on different platforms like smartphones.}
    \sethlcolor{pink!100}
    \hl{Our paper presents an efficient framework for creating high resolution stereo pictures from 2D videos using only image-level information.}
    \sethlcolor{pink!60}
    \hl{Even in challenging conditions like low frame rate, poor lighting, and occlusion propagation, our approach can produce detailed views.}
    \sethlcolor{pink!60}
    \hl{Additionally, our method minimizes computational cost and reduces noise during preprocessing to maintain high quality.}
    \sethlcolor{pink!40}
    \hl{It allows for regular interval image fitting without significant loss in accuracy.}
    \sethlcolor{pink!40}
    \hl{It also provides real-time camera pose updates, as demonstrated by conducting human arm position estimation over a webcam stream with accuracies comparable to other methods.}
    \sethlcolor{pink!15}
    \hl{In addition, full colour images were produced within 60 ms on device with limited processing power e.g., Galaxy S5 smartphone equipped with Adreno 320 GPU.}
    \sethlcolor{pink!15}
    \hl{Finally most importantly our technique achieves good performance in terms of energy efficiency allowing to use a single thread of CPU with minimal memory requirement.} \\
    \midrule
    \midrule
    \textbf{Raw Text}: 
    \sethlcolor{cyan!60}
    \hl{Currently, the issue of offensive content on social media is a significant concern.}
    \sethlcolor{cyan!100}
    \hl{It is crucial to have a system that can automatically identify offensive language.}
    \sethlcolor{cyan!100}
    \hl{In this research paper, we develop a cloud-based offensive language detection system.}
    \sethlcolor{cyan!40}
    \hl{By utilizing data from users' web browsing history obtained through Google Analytics, we analyze CSS and generate text messages for display in ads sent via the Google AdWords channel.}
    \sethlcolor{cyan!15}
    \hl{Furthermore, we compare these messages with other harmful links detected through Google Analytics to identify potentially dangerous advertising campaigns.}
    \sethlcolor{cyan!60}
    \hl{These campaigns are designed to target individuals who download advertisements from websites offering free cookies or torrent sites.} \\
    \midrule
    \textbf{MFD Detection Result}: 
    \sethlcolor{pink!60}
    \hl{Currently, the issue of offensive content on social media is a significant concern.}
    \sethlcolor{pink!60}
    \hl{It is crucial to have a system that can automatically identify offensive language.}
    \sethlcolor{pink!100}
    \hl{In this research paper, we develop a cloud-based offensive language detection system.}
    \sethlcolor{pink!40}
    \hl{By utilizing data from users' web browsing history obtained through Google Analytics, we analyze CSS and generate text messages for display in ads sent via the Google AdWords channel.}
    \sethlcolor{pink!15}
    \hl{Furthermore, we compare these messages with other harmful links detected through Google Analytics to identify potentially dangerous advertising campaigns.}
    \sethlcolor{pink!60}
    \hl{These campaigns are designed to target individuals who download advertisements from websites offering free cookies or torrent sites.} \\
    \bottomrule
    \end{tabular}}
    \begin{tablenotes}
        \footnotesize
        \item \textit{Notes}: Blue represents the ground truth, and pink indicates the predicted values. Darker shades correspond to higher levels of LLM \\ involvement, which is divided into four distinct levels. A threshold of 5\% is used, below which the text is considered free of LLM \\ involvement.
    \end{tablenotes}
  \end{threeparttable}
  \label{tab7}%
\end{table*}%

\subsection{Discussion}

The experimental results demonstrate the effectiveness and robustness of our proposed framework for detecting LLM involvement in LLM-generated text across multiple evaluation settings. Through comparative analysis with state-of-the-art models, our method consistently achieved superior performance. The integration of multi-level features (low, high, and deep) proved to be essential, as shown in the ablation studies, where combining low-level and deep-level features notably improved the detection accuracy. Our model’s performance across different LLMs and encoders further validates its versatility and adaptability. Notably, T5's architecture and Llama-3.1-8B-Instruct provided the most balanced performance across metrics, underscoring the effectiveness of these models in capturing subtle manipulations typically found in LLM-generated text. The generalization performance study also confirms that our method can effectively adapt to out-of-domain data, maintaining strong performance even when applied to text that has been stylistically altered by advanced models like GPT-4.

These results suggest several key insights. First, the integration of multi-level linguistic features is critical for improving the detection of LLM-generated text, as different features capture distinct aspects of linguistic manipulation. Second, the choice of encoder or LLM significantly impacts the performance, with models like T5 offering more versatility due to their sequence-to-sequence structure. Lastly, while higher accuracy thresholds can lead to more precise detection in certain linguistic dimensions, a balanced threshold (such as 0.5) ensures stable performance without risking overfitting. Future work should focus on further optimizing these thresholds for different tasks and exploring how additional linguistic features, such as pragmatic or discourse-level information, might enhance model performance. Moreover, expanding the evaluation to more diverse and complex datasets could provide deeper insights into the model’s robustness across various writing styles and domains.

\section{Conclusion and Future Work}\label{sec5}

In this paper, we propose a novel \textbf{M}ulti-level \textbf{F}ine-grained \textbf{D}etection (\textbf{MFD}) framework that effectively addresses the growing challenges in detecting LLM-generated text in academic writing. By incorporating a multi-level analysis of statistical, semantic, and linguistic features at the sentence level, our approach not only enhances detection accuracy but also offers a more granular evaluation of LLM involvement. Moreover, to ensure consistent performance across diverse text variations, our framework leverages contrastive learning on both original and manipulated LLM-generated text. This design aims to enhance robustness, making it more resilient to common evasion tactics. Extensive experiments conducted on the public PASTED dataset demonstrate that our MFD model outperforms existing methods, achieving an MAE of 0.1346 and an accuracy of 88.56\%. Additionally, the model maintains consistent performance on MP and OOD-GPT4 datasets, validating its strong generalization capabilities across various types of LLM-generated text. Our approach makes a significant contribution to the field of academic integrity by providing a comprehensive and scalable solution for the precise identification of LLM-generated text, which is critical in preserving the authenticity and credibility of scholarly research in the evolving landscape of LLM-driven text generation.

Looking ahead, several important directions for future research emerge. One promising path is to extend the framework's capabilities to a cross-linguistic context, enabling the detection of LLM-generated text across multiple languages, which is crucial given the global nature of academic discourse. Additionally, exploring adaptive learning techniques could further strengthen the model’s resilience to advancements in LLM technologies and sophisticated adversarial attempts to bypass detection systems. Another valuable area for development lies in integrating real-time detection functionalities into academic writing tools, offering immediate safeguards to uphold the integrity and originality of scholarly output. By advancing these areas, the framework can continue to lead efforts in detecting LLM-generated text and protecting academic integrity in an evolving digital landscape.


\bibliographystyle{elsarticle-num}

\bibliography{cas-refs}

\begin{thebibliography}{10}
\expandafter\ifx\csname url\endcsname\relax
  \def\url#1{\texttt{#1}}\fi
\expandafter\ifx\csname urlprefix\endcsname\relax\def\urlprefix{URL }\fi
\expandafter\ifx\csname href\endcsname\relax
  \def\href#1#2{#2} \def\path#1{#1}\fi

\bibitem{2023co}
M.~Jakesch, A.~Bhat, D.~Buschek, L.~Zalmanson, M.~Naaman, Co-writing with opinionated language models affects users’ views, in: Proceedings of the 2023 CHI conference on human factors in computing systems, 2023, pp. 1--15.

\bibitem{2023bookgptli}
Z.~Li, Y.~Chen, X.~Zhang, X.~Liang, Bookgpt: A general framework for book recommendation empowered by large language model, Electronics 12~(22) (2023) 4654.

\bibitem{2024evaluation}
P.~Hager, F.~Jungmann, R.~Holland, K.~Bhagat, I.~Hubrecht, M.~Knauer, J.~Vielhauer, M.~Makowski, R.~Braren, G.~Kaissis, et~al., Evaluation and mitigation of the limitations of large language models in clinical decision-making, Nature medicine (2024) 1--10.

\bibitem{2024tree}
S.~Yao, D.~Yu, J.~Zhao, I.~Shafran, T.~Griffiths, Y.~Cao, K.~Narasimhan, Tree of thoughts: Deliberate problem solving with large language models, Advances in Neural Information Processing Systems 36 (2024).

\bibitem{2023written}
B.~D. Lund, T.~Wang, N.~R. Mannuru, B.~Nie, S.~Shimray, Z.~Wang, Chatgpt and a new academic reality: Artificial intelligence-written research papers and the ethics of the large language models in scholarly publishing, Journal of the Association for Information Science and Technology 74~(5) (2023) 570--581.

\bibitem{2023science}
A.~Birhane, A.~Kasirzadeh, D.~Leslie, S.~Wachter, Science in the age of large language models, Nature Reviews Physics 5~(5) (2023) 277--280.

\bibitem{2023empowering}
S.~Moore, R.~Tong, A.~Singh, Z.~Liu, X.~Hu, Y.~Lu, J.~Liang, C.~Cao, H.~Khosravi, P.~Denny, et~al., Empowering education with llms-the next-gen interface and content generation, in: International Conference on Artificial Intelligence in Education, Springer, 2023, pp. 32--37.

\bibitem{2024bias}
S.~Dai, C.~Xu, S.~Xu, L.~Pang, Z.~Dong, J.~Xu, Bias and unfairness in information retrieval systems: New challenges in the llm era, in: Proceedings of the 30th ACM SIGKDD Conference on Knowledge Discovery and Data Mining, 2024, pp. 6437--6447.

\bibitem{2024exploring}
G.~F. Almeida, J.~L. Nunes, N.~Engelmann, A.~Wiegmann, M.~de~Ara{\'u}jo, Exploring the psychology of llms’ moral and legal reasoning, Artificial Intelligence 333 (2024) 104145.

\bibitem{2023overview}
M.~Kayyali, An overview of quality assurance in higher education: Concepts and frameworks, International Journal of Management, Sciences, Innovation, and Technology (IJMSIT) 4~(2) (2023) 01--04.

\bibitem{2024practical}
L.~Yan, L.~Sha, L.~Zhao, Y.~Li, R.~Martinez-Maldonado, G.~Chen, X.~Li, Y.~Jin, D.~Ga{\v{s}}evi{\'c}, Practical and ethical challenges of large language models in education: A systematic scoping review, British Journal of Educational Technology 55~(1) (2024) 90--112.

\bibitem{2024llm}
Q.~Zhang, C.~Gao, D.~Chen, Y.~Huang, Y.~Huang, Z.~Sun, S.~Zhang, W.~Li, Z.~Fu, Y.~Wan, et~al., Llm-as-a-coauthor: Can mixed human-written and machine-generated text be detected?, in: Findings of the Association for Computational Linguistics: NAACL 2024, 2024, pp. 409--436.

\bibitem{2024ACMsurvey}
H.~Lai, M.~Nissim, A survey on automatic generation of figurative language: From rule-based systems to large language models, ACM Computing Surveys 56~(10) (2024) 1--34.

\bibitem{2024authorship}
B.~Huang, C.~Chen, K.~Shu, Authorship attribution in the era of llms: Problems, methodologies, and challenges, arXiv preprint arXiv:2408.08946 (2024).

\bibitem{2023Frontiers}
L.~De~Angelis, F.~Baglivo, G.~Arzilli, G.~P. Privitera, P.~Ferragina, A.~E. Tozzi, C.~Rizzo, Chatgpt and the rise of large language models: the new ai-driven infodemic threat in public health, Frontiers in public health 11 (2023) 1166120.

\bibitem{2024navigating}
Y.~Zhou, B.~He, L.~Sun, Navigating the shadows: Unveiling effective disturbances for modern ai content detectors, arXiv preprint arXiv:2406.08922 (2024).

\bibitem{2024hidding}
X.~Peng, Y.~Zhou, B.~He, L.~Sun, Y.~Sun, Hidding the ghostwriters: An adversarial evaluation of ai-generated student essay detection, arXiv preprint arXiv:2402.00412 (2024).

\bibitem{2024survey}
Y.~Chang, X.~Wang, J.~Wang, Y.~Wu, L.~Yang, K.~Zhu, H.~Chen, X.~Yi, C.~Wang, Y.~Wang, et~al., A survey on evaluation of large language models, ACM Transactions on Intelligent Systems and Technology 15~(3) (2024) 1--45.

\bibitem{2023canchen}
C.~Chen, K.~Shu, Can llm-generated misinformation be detected?, arXiv preprint arXiv:2309.13788 (2023).

\bibitem{2023academicperkins}
M.~Perkins, Academic integrity considerations of ai large language models in the post-pandemic era: Chatgpt and beyond, Journal of University Teaching and Learning Practice 20~(2) (2023).

\bibitem{2021algorithmicbirhane}
A.~Birhane, Algorithmic injustice: a relational ethics approach, Patterns 2~(2) (2021).

\bibitem{2019gltr}
S.~Gehrmann, H.~Strobelt, A.~M. Rush, Gltr: Statistical detection and visualization of generated text, arXiv preprint arXiv:1906.04043 (2019).

\bibitem{2019release}
I.~Solaiman, M.~Brundage, J.~Clark, A.~Askell, A.~Herbert-Voss, J.~Wu, A.~Radford, G.~Krueger, J.~W. Kim, S.~Kreps, et~al., Release strategies and the social impacts of language models, arXiv preprint arXiv:1908.09203 (2019).

\bibitem{2023detectgpt}
E.~Mitchell, Y.~Lee, A.~Khazatsky, C.~D. Manning, C.~Finn, Detectgpt: Zero-shot machine-generated text detection using probability curvature, in: International Conference on Machine Learning, PMLR, 2023, pp. 24950--24962.

\bibitem{2023multiscale}
Y.~Tian, H.~Chen, X.~Wang, Z.~Bai, Q.~Zhang, R.~Li, C.~Xu, Y.~Wang, Multiscale positive-unlabeled detection of ai-generated texts, arXiv preprint arXiv:2305.18149 (2023).

\bibitem{2023close}
B.~Guo, X.~Zhang, Z.~Wang, M.~Jiang, J.~Nie, Y.~Ding, J.~Yue, Y.~Wu, How close is chatgpt to human experts? comparison corpus, evaluation, and detection, arXiv preprint arXiv:2301.07597 (2023).

\bibitem{2023seqxgpt}
P.~Wang, L.~Li, K.~Ren, B.~Jiang, D.~Zhang, X.~Qiu, Seqxgpt: Sentence-level ai-generated text detection, arXiv preprint arXiv:2310.08903 (2023).

\bibitem{2023coco}
X.~Liu, Z.~Zhang, Y.~Wang, H.~Pu, Y.~Lan, C.~Shen, Coco: Coherence-enhanced machine-generated text detection under low resource with contrastive learning, in: Proceedings of the 2023 Conference on Empirical Methods in Natural Language Processing, 2023, pp. 16167--16188.

\bibitem{2023multiscalelu}
S.~Lu, Y.~Ding, M.~Liu, Z.~Yin, L.~Yin, W.~Zheng, Multiscale feature extraction and fusion of image and text in vqa, International Journal of Computational Intelligence Systems 16~(1) (2023) 54.

\bibitem{2023multiislam}
M.~M. Islam, S.~Nooruddin, F.~Karray, G.~Muhammad, Multi-level feature fusion for multimodal human activity recognition in internet of healthcare things, Information Fusion 94 (2023) 17--31.

\bibitem{2024parttian}
Y.~Tian, R.~Yue, D.~Wang, J.~Liu, X.~Liang, Part-of-speech-and syntactic-aware graph convolutional network for aspect-level sentiment classification, Multimedia Tools and Applications 83~(10) (2024) 28793--28806.

\bibitem{2023neuralkazanina}
N.~Kazanina, A.~Tavano, What neural oscillations can and cannot do for syntactic structure building, Nature Reviews Neuroscience 24~(2) (2023) 113--128.

\bibitem{2023contrastingmunoz}
A.~Mu{\~n}oz-Ortiz, C.~G{\'o}mez-Rodr{\'\i}guez, D.~Vilares, Contrasting linguistic patterns in human and llm-generated text, arXiv preprint arXiv:2308.09067 (2023).

\bibitem{2023gptachiam}
J.~Achiam, S.~Adler, S.~Agarwal, L.~Ahmad, I.~Akkaya, F.~L. Aleman, D.~Almeida, J.~Altenschmidt, S.~Altman, S.~Anadkat, et~al., Gpt-4 technical report, arXiv preprint arXiv:2303.08774 (2023).

\bibitem{2023exploring}
A.~R. Malik, Y.~Pratiwi, K.~Andajani, I.~W. Numertayasa, S.~Suharti, A.~Darwis, et~al., Exploring artificial intelligence in academic essay: higher education student's perspective, International Journal of Educational Research Open 5 (2023) 100296.

\bibitem{2024using}
M.~Khalifa, M.~Albadawy, Using artificial intelligence in academic writing and research: An essential productivity tool, Computer Methods and Programs in Biomedicine Update (2024) 100145.

\bibitem{2024mapping}
W.~Liang, Y.~Zhang, Z.~Wu, H.~Lepp, W.~Ji, X.~Zhao, H.~Cao, S.~Liu, S.~He, Z.~Huang, et~al., Mapping the increasing use of llms in scientific papers, arXiv preprint arXiv:2404.01268 (2024).

\bibitem{2024generative}
M.~Taiye, C.~High, J.~Velander, K.~Matar, R.~Okmanis, M.~Milrad, Generative ai-enhanced academic writing: A stakeholder-centric approach for the design and development of chat4isp-ai, in: Proceedings of the 39th ACM/SIGAPP Symposium on Applied Computing, 2024, pp. 74--80.

\bibitem{2024utilizing}
M.~Safrai, K.~E. Orwig, Utilizing artificial intelligence in academic writing: an in-depth evaluation of a scientific review on fertility preservation written by chatgpt-4, Journal of Assisted Reproduction and Genetics (2024) 1--10.

\bibitem{2024chatgptmogavi}
R.~H. Mogavi, C.~Deng, J.~J. Kim, P.~Zhou, Y.~D. Kwon, A.~H.~S. Metwally, A.~Tlili, S.~Bassanelli, A.~Bucchiarone, S.~Gujar, et~al., Chatgpt in education: A blessing or a curse? a qualitative study exploring early adopters’ utilization and perceptions, Computers in Human Behavior: Artificial Humans 2~(1) (2024) 100027.

\bibitem{2024llmszhang}
J.~Zhang, H.~Bu, H.~Wen, Y.~Chen, L.~Li, H.~Zhu, When llms meet cybersecurity: A systematic literature review, arXiv preprint arXiv:2405.03644 (2024).

\bibitem{2024usepatel}
E.~A. Patel, L.~Fleischer, P.~Filip, M.~Eggerstedt, M.~Hutz, E.~Michaelides, P.~S. Batra, B.~A. Tajudeen, The use of artificial intelligence to improve readability of otolaryngology patient education materials, Otolaryngology--Head and Neck Surgery (2024).

\bibitem{2024advancingjeong}
Y.~Jeong, J.-J. Song, J.~Yang, S.~Kang, Advancing tinnitus therapeutics: Gpt-2 driven clustering analysis of cognitive behavioral therapy sessions and google t5-based predictive modeling for thi score assessment, IEEE Access (2024).

\bibitem{2024andkhurana}
A.~Khurana, H.~Subramonyam, P.~K. Chilana, Why and when llm-based assistants can go wrong: Investigating the effectiveness of prompt-based interactions for software help-seeking, in: Proceedings of the 29th International Conference on Intelligent User Interfaces, 2024, pp. 288--303.

\bibitem{2024creatingguerin}
C.~Guerin, C.~Aitchison, S.~Carter, Creating, Managing, and Editing Multi-authored Publications: A Guide for Scholars, Taylor \& Francis, 2024.

\bibitem{2024outfoxkoike}
R.~Koike, M.~Kaneko, N.~Okazaki, Outfox: Llm-generated essay detection through in-context learning with adversarially generated examples, in: Proceedings of the AAAI Conference on Artificial Intelligence, Vol.~38, 2024, pp. 21258--21266.

\bibitem{2024stumbling}
Y.~Wang, S.~Feng, A.~B. Hou, X.~Pu, C.~Shen, X.~Liu, Y.~Tsvetkov, T.~He, Stumbling blocks: Stress testing the robustness of machine-generated text detectors under attacks, arXiv preprint arXiv:2402.11638 (2024).

\bibitem{2023largekaranikolas}
N.~Karanikolas, E.~Manga, N.~Samaridi, E.~Tousidou, M.~Vassilakopoulos, Large language models versus natural language understanding and generation, in: Proceedings of the 27th Pan-Hellenic Conference on Progress in Computing and Informatics, 2023, pp. 278--290.

\bibitem{2024formohmer}
X.~Ohmer, E.~Bruni, D.~Hupkes, From form (s) to meaning: Probing the semantic depths of language models using multisense consistency, Computational Linguistics (2024) 1--51.

\bibitem{2024spottingli}
Y.~Li, Z.~Wang, L.~Cui, W.~Bi, S.~Shi, Y.~Zhang, Spotting ai's touch: Identifying llm-paraphrased spans in text, arXiv preprint arXiv:2405.12689 (2024).

\bibitem{li2023origin}
L.~Li, P.~Wang, K.~Ren, T.~Sun, X.~Qiu, Origin tracing and detecting of llms, arXiv preprint arXiv:2304.14072 (2023).

\bibitem{2024machinezhang}
Z.~Zhang, W.~Qin, B.~A. Plummer, Machine-generated text localization, arXiv preprint arXiv:2402.11744 (2024).

\end{thebibliography}



\end{document}